\documentclass[a4paper,twoside]{article}

\usepackage{epsfig}
\usepackage{subcaption}
\usepackage{calc}
\usepackage{amssymb}
\usepackage{amstext}
\usepackage{amsmath}
\usepackage{amsthm}
\usepackage{multicol}
\usepackage{pslatex}
\usepackage{apalike}
\usepackage{enumitem} 
\usepackage{SCITEPRESS}     

\begin{document}

\title{Uncertainty-Based Out-of-Distribution Classification in~Deep~Reinforcement~Learning}

\author{\authorname{Andreas Sedlmeier\sup{1}, Thomas Gabor\sup{1}, Thomy Phan\sup{1},\\Lenz Belzner\sup{2} and Claudia Linnhoff-Popien\sup{1}}
  \affiliation{\sup{1}LMU Munich, Munich, Germany}
  \affiliation{\sup{2}MaibornWolff, Munich, Germany}
  \email{andreas.sedlmeier@ifi.lmu.de}
}


\abstract{
  Robustness to out-of-distribution (OOD) data is an important goal in building reliable machine learning systems.
  Especially in autonomous systems, wrong predictions for OOD inputs can cause safety critical situations.
  As a first step towards a solution, we consider the problem of detecting such data in a value-based deep reinforcement learning (RL) setting.
  Modelling this problem as a one-class classification problem, we propose a framework for uncertainty-based OOD classification: UBOOD.
  It is based on the effect that an agent's epistemic uncertainty is reduced for situations encountered during training (in-distribution), and thus lower than for unencountered (OOD) situations.
  Being agnostic towards the approach used for estimating epistemic uncertainty, combinations with different uncertainty estimation methods, e.g. approximate Bayesian inference methods or ensembling techniques are possible.
  We further present a first viable solution for calculating a dynamic classification threshold, based on the uncertainty distribution of the training data.
  Evaluation shows that the framework produces reliable classification results when combined with ensemble-based estimators,
  while the combination with concrete dropout-based estimators fails to reliably detect OOD situations.
  In summary, UBOOD presents a viable approach for OOD classification in deep RL settings by leveraging the epistemic uncertainty of the agent's value function.
}

\onecolumn \maketitle \normalsize \setcounter{footnote}{0} \vfill

\section{\uppercase{Introduction}}
\label{sec:introduction}

\noindent One of the main impediments to the deployment of autonomous machine learning systems in the real world
is the difficulty to show that the system will continue to reliably execute beneficial actions in all the situations it encounters in production use.
One of the possible reasons for failure is so called out-of-distribution (OOD) data, i.e. data which deviates substantially from the data encountered during training.
As the fundamental problem of limited training data seems unsolvable for most cases, especially in sequential decision making tasks like reinforcement learning (RL),
a possible first step towards a solution is to detect and report the occurrence of OOD data.
This can prevent silent and possibly safety critical failures of the machine learning system (caused by wrong predictions which lead to the execution of unfavorable actions), for example by handing control over to a human supervisor \cite{amodei16}. 
Recently, several different approaches were proposed that try to detect OOD samples in classification tasks \cite{hendrycks16,liang17}, or perform anomaly detection via generative models \cite{schlegl17}.
While these methods show promising results in the evaluated classification tasks, we are not aware of applications to value-based RL settings where non-stationary regression targets are present.
Thus, our research aims to provide a first step towards developing and evaluating suitable OOD detection methods that are applicable to changing environments in sequential decision making tasks.
We model the OOD-detection problem as a one-class classification problem with the two classes: in-distribution and out-of-distribution.
Having framed the problem this way, we propose a framework for uncertainty-based OOD classification: UBOOD.
It is based on the effect that epistemic uncertainty in the agent's chosen actions is reduced for situations encountered during training (in-distribution), and is thus lower than for unencountered (OOD) situations.
The framework itself is agnostic towards the approach used for estimating epistemic uncertainty.
Thus, it is possible to use e.g. approximate Bayesian inference methods or ensembling techniques.
In order to evaluate the performance of any OOD classifier in a RL setting, modifiable environments which can generate OOD samples are needed.
Due to a lack of publicly available RL environments that allow systematic modification, we developed two different environments: one using a gridworld-style discrete state-space, the other using a continuous state-space.
Both allow modifications of increasing strength (and consequently produce OOD samples of increasing strength) after the training process.
We empirically evaluated the performance of the UBOOD framework with different uncertainty estimation methods on these environments.
Evaluation results show that the framework produces reliable OOD classification results when combined with ensemble-based estimators,
while the combination with concrete dropout-based estimators fails to capture increased uncertainty in the OOD situations.
Ensemble-based approaches also show increasing classification accuracy, the \textit{stronger} the OOD samples are (i.e. the more the environments differ from training)
and increasing uncertainty is inversely related with the agent's achieved return.

\section{\uppercase{Basics}}

\subsection{Uncertainty}
When viewed from a statistical perspective, uncertainty arises whenever the outcome of a random variable cannot be known with certainty.
Uncertainty measures can then be understood to describe how random the outcome of such a random variable is.
This ``amount of randomness" is described by the dispersion of the random variable's probability distribution, i.e. how stretched or squeezed the probability distribution is.
Measures of this dispersion are e.g. the probability distribution's variance or standard deviation. \cite{bishop2006pattern}

\subsubsection{Uncertainty Estimation}
\begin{figure}[t]
  \includegraphics[width=.48\textwidth]{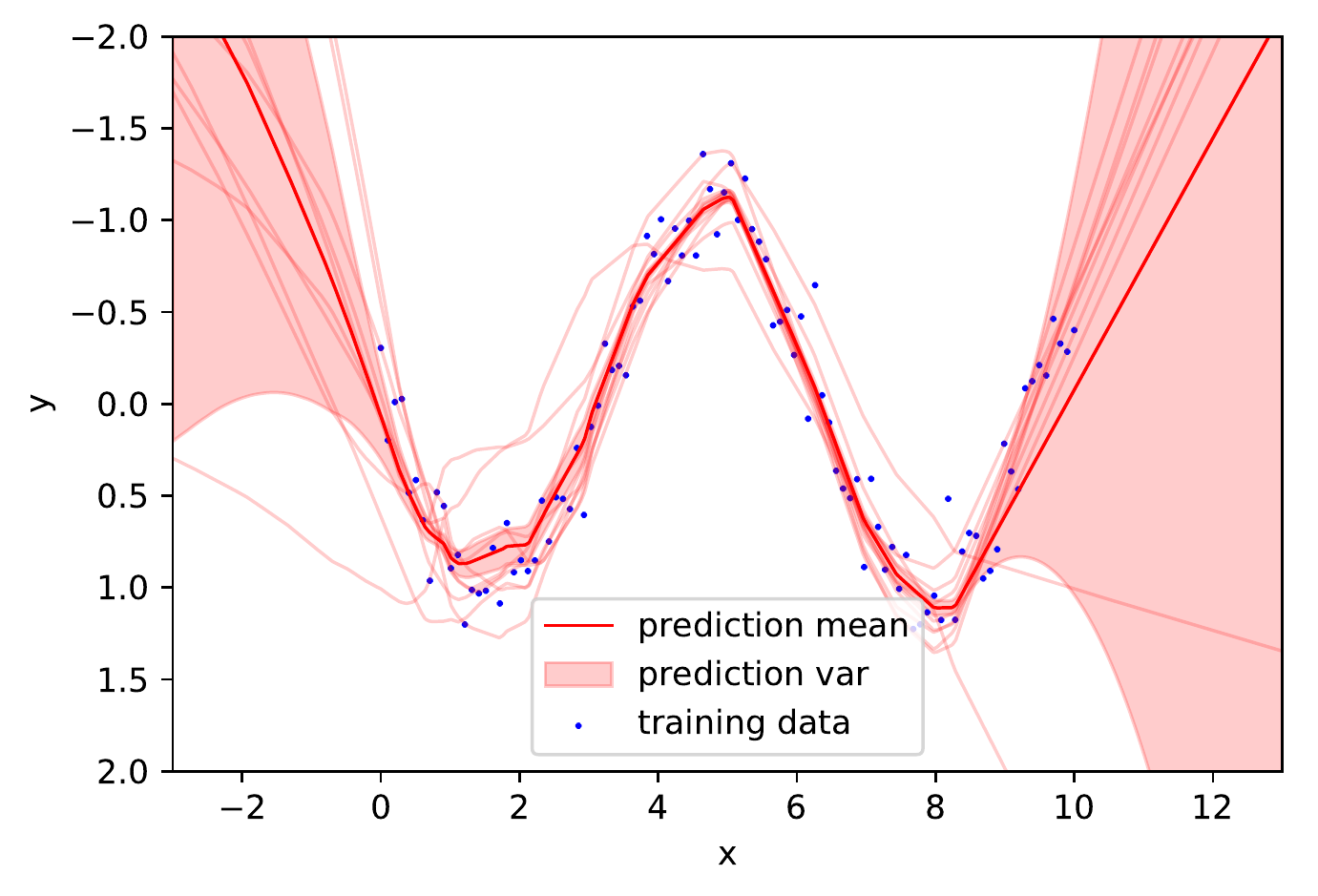}
  \caption{Example regression of a 1-D toy-dataset showing the predictions of a Bootstrap ensemble (see Section \ref{subsec_epis_unc_est_approaches}) of size $10$. Blue dots represent the training data. Thin red lines show the individual ensemble predictions, while the thick red line represents the mean of the predictions. The variance of the individual predictions can be interpreted as epistemic uncertainty.}
  \label{fig:bootstrap_var}
\end{figure}

In the context of this work, we are interested in the uncertainty of a neural network's prediction, which in a value-based deep RL setting 
is the certainty that an agent's chosen action is optimal in the given situation.
Different approaches exist that make it possible to estimate this uncertainty.
Ensemble techniques for example aggregate the predictions of multiple networks,
often trained on different versions of the data, and interpret the variance of the individual predictions as the uncertainty \cite{bootstrappedDQN16,lakshminarayanan17}.
An example of this approach can be seen in Figure \ref{fig:bootstrap_var}, which shows the individual predictions of a Bootstrap ensemble as well as their mean and variance.
These and other methods applicable to deep neural networks will be presented in more detail in Section \ref{subsec:uncertainty_in_dl}.
Besides the various ways of measuring uncertainty, it is equally important to differentiate the different sources of uncertainty.

\subsubsection{Aleatoric Uncertainty}
Aleatoric uncertainty models the inherent stochasticity in the system, i.e. no amount of data can explain the observed stochasticity.
In other words, the uncertainty cannot be reduced by capturing more data.
A reason for this might be that certain features that would be needed to explain the behaviour of the system are not part of the collected data.
E.g. consider trying to model the distance different cars travel on a highway in a certain amount of time, without measuring their speed.
If the speed is not part of the collected data, the \textit{randomness} in the measured distances cannot be explained.
It is also possible that the uncertainty is a fundamental property of the measured system, as is the case when dealing with quantum mechanics.
As such, aleatoric uncertainty cannot be reduced, irrespective of how much data is collected.

\subsubsection{Epistemic Uncertainty}
Epistemic uncertainty by contrast arises out of a lack of sufficient data to exactly infer the underlying system's data generating function.
In this case, the features available in the data do in principle allow the explanation of the behaviour of the system.
In the previous example, this would e.g. be the case if both time and speed are measured, but so far only cars traveling at the same speed had been observed.
The uncertainty caused by the effect of different speed in this case is epistemic, as collecting more data could allow for a correct inference of the system's behaviour
and consequently the reduction of the uncertainty.

\subsection{Markov Decision Processes}
We base our problem formulation on Markov decision processes (MDPs) \cite{puterman2014markov}. MDPs are defined by tuples: $\mathcal{M} = \langle \mathcal{S}, \mathcal{A}, \mathcal{P}, \mathcal{R} \rangle$.
$\mathcal{S}$ is a (finite) set of states; $s_t \in \mathcal{S}$ being the state of the MDP at time step $t$. $\mathcal{A}$ is the (finite) set of actions; $a_t \in \mathcal{A}$ is the action the MDP takes at step $t$. $\mathcal{P}(s_{t+1}|s_t,a_t)$ defines the transition probability function; a transition occurs by executing action $a_t$ in state $s_t$. The resulting next state $s_{t+1}$ is determined based on $\mathcal{P}$. In this paper we focus on deterministic domains represented by deterministic MDPs, so $\mathcal{P}(s_{t+1}|s_t, a_t) \in \{0, 1\}$. Finally, $\mathcal{R}(s_t,a_t)$ is the scalar reward; for this paper we assume that  $\mathcal{R}(s_t,a_t) \in \mathbb{R}$.

Goal of the problem is to find a policy $\pi : \mathcal{S} \rightarrow \mathcal{A}$ in the space of all possible policies $\Pi$, which maximizes the expectation of return $G_t$ at state $s_t$ over a potentially infinite horizon:
\begin{equation}
  G_{t} = \sum_{k=0}^{\infty} \gamma^{k} \cdot \mathcal{R}(s_{t+k}, a_{t+k})
\end{equation}
where $\gamma \in [0,1]$ is the discount factor.

\subsection{Reinforcement Learning}

In order to search the policy space $\Pi$, we consider model-free reinforcement learning (RL). In this setting, an agent interacts with an environment defined as an MDP $\mathcal{M}$ by executing a sequence of actions $a_t \in \mathcal{A}, t = 0, 1, ...$~\cite{sutton1998introduction}.
In the fully observable case of RL, the agent knows its current state $s_t$ and the action space $\mathcal{A}$, but not the effect of executing $a_t$ in $s_t$, i.e., $\mathcal{P}(s_{t+1} | s_t, a_t)$ and $\mathcal{R}(s_t, a_t)$.
In order to find the optimal policy $\pi^*$, we focus on Q-Learning \cite{watkins1989learning}, a commonly used value-based approach.
It is named for the action-value function $Q^{\pi}: \mathcal{S} \times \mathcal{A} \rightarrow \mathbb{R}, \pi \in \Pi$,
which describes the expected return $Q^\pi(s_t, a_t)$ when taking action $a_t$ in state $s_t$ and then following policy $\pi$ for all states $s_{t+1}, s_{t+2}, ...$ afterwards.

The optimal action-value function $Q^{*}$ of policy $\pi^*$ is any action-value function that yields higher accumulated rewards than all other action-value functions, i.e., $Q^{*}(s_t,a_t) \geq Q^{\pi}(s_t,a_t)\;\forall \pi \in \Pi$.
Q-Learning aims to approximate $Q^*$ by starting from an initial guess for $Q$, which is then updated via
\begin{align}
  Q(s_t, a_t) \leftarrow \; & Q(s_t, a_t) + \nonumber \\&\alpha [r_t + \gamma \max\limits_a Q(s_{t+1}, a) - Q(s_t, a_t)]
\end{align}

It uses experience samples of the form $e_t = (s_t, a_t, s_{t+1}, r_t)$, where $r_t$ is the reward earned at time step $t$, i.e., by executing action $a_t$ when in state $s_t$.
The learning rate $\alpha$ is a setup-specific parameter.
The set of all experience samples taken at time steps $t_1, ..., t_m$ for some training limit $m$ is called the training set $\mathcal{T} = \{e_{t_1}, ..., e_{t_m}\}$.

The learned action-value function $Q$ converges to the optimal action-value function $Q^*$, which then implies an optimal policy $\pi^*(s_t) = \arg\!\max_a Q(s_t, a)$.

In high-dimensional settings or when learning in continuous state-spaces, it is common to use parameterized function approximators like neural networks to approximate the action-value function:
$Q(s_t, a_t;\theta) \approx Q^*(s_t, a_t)$ with $\theta$ specifying the weights of the neural network.
When using a deep neural network as the function approximator, this approach is called deep reinforcement learning. \cite{mnih2015human}

\section{\uppercase{Related Work}}

\subsection{Uncertainty in Deep Learning}
\label{subsec:uncertainty_in_dl}
When dealing with uncertainty, a systematic way is via Bayesian inference.
Its combination with neural networks in the form of Bayesian neural networks is realised by placing a probability distribution over the weight-values of the network \cite{mackay92}.
As calculating the exact Bayesian posterior quickly becomes computationally intractable for deep models, a popular solution are approximate inference methods
\cite{graves11,blundell15,gal16dropout,hernandez16,li17,galConcrete17}.
Another option is the construction of model ensembles, e.g., based on the idea of the statistical bootstrap \cite{efron92}.
The resulting distribution of the ensemble predictions can then be used to approximate the uncertainty \cite{bootstrappedDQN16,lakshminarayanan17}.

Both approaches have been used for tasks as diverse as machine vision \cite{whatUncertainties17} or disease detection \cite{leibig17}.
In the field of decision making, uncertainty is used to implicitly guide exploration, e.g by creating an ensemble of models \cite{bootstrappedDQN16},
or for learning safety predictors, e.g. predicting the probability of a collision \cite{kahn2017uncertainty}.
Recently, a distributional approach to RL \cite{bellemare2017distributional} was proposed which tries to learn the value distribution of a RL environment.
Although this approach also models uncertainty, its goal of estimating the distribution of values is different from the work at hand,
which tries to detect epistemic uncertainty, i.e. uncertainty in the model itself.

\subsection{OOD and Novelty Detection}
For the case of low-dimensional feature spaces, OOD detection (also called novelty detection) is a well-researched problem.
For a survey on the topic, see e.g. \cite{pimentel14}, who distinguish between probabilistic, distance-based, reconstruction-based, domain-based and information theoretic methods.
During the last years, several new methods based on deep neural networks were proposed for high-dimensional cases, mostly focusing on classification tasks, e.g. image classification.
\cite{hendrycks16} propose a baseline for detecting OOD examples in neural networks, based on the predicted class probabilities of a softmax classifier.
\cite{liang17} improve upon this baseline by using temperature scaling and by adding perturbations to the input.
\cite{li17} evaluate the performance of a proposed alpha-divergence-based variational inference technique in an image classification task of adversarial examples.
This can be understood as a form of OOD detection, as the generated adversarial examples lie outside of the training image manifold and consequently far from the training data.
The authors report increased epistemic uncertainty, confirming the viability of their approach for the detection of adversarial image examples. 
The basic idea of this uncertainty-based approach is closely related to our proposed method, but no evaluation of the performance in a RL setting with non-stationary regression targets was performed.
To the best our knowledge, none of the previously mentioned methods were evaluated regarding the epistemic uncertainty detection performance in a RL setting.

\section{\uppercase{UBOOD: Uncertainty-Based Out-of-Distribution Classification}}\label{sec:u-bood}
\noindent In this paper we propose UBOOD, an uncertainty-based OOD-classifier that can be employed in value-based deep reinforcement learning settings.
It is based on the reducibility of epistemic uncertainty in the action-value function approximation.

As previously described, epistemic uncertainty arises out of a lack of sufficient data to exactly infer the underlying system's data generating function.
As such, it tends to be higher in areas of low data density.
\cite{qazaz96}, who in turn refers to \cite{bishop94} for the initial conjecture, 
showed that the epistemic uncertainty $\sigma_{epis}(x)$ is approximately inversely proportional to the density $p(x)$ of the input data,
for the case of generalized linear regression models as well as multi-layer neural networks:
\begin{equation}\label{eq_quazaz}
  \sigma_{epis}(x) \propto p^{-1}(x)
\end{equation}

This also forms the basis of our approach: to use this inverse relation between epistemic uncertainty and data density in order to differentiate in- from out-of-distribution samples.

We define $U_Q: \mathcal{S} \times \mathcal{A} \rightarrow \mathbb{R}$ as the epistemic uncertainty function of a given Q-function approximation $Q$.
If a suitable method for epistemic uncertainty estimation for deep neural networks is applied,
the process of training the agent reduces $U_Q(s, a)$ for those state-action tuples $(s,a) \in \mathbb{I}$ that were used for training,
i.e., there exists a successor state $s'$ and a reward $r$ so that $(s, a, s', r) \in \mathcal{I}$.
$\mathbb{I}$ consequently defines the set of in-distribution data.
By contrast, state-action tuples that were not encountered during training
i.e. $(s,a) \not\in \mathbb{I}$ define the set of out-of-distribution data $\mathbb{O}$.
The epistemic uncertainty of these state-action tuples is not reduced during training.
Thus, epistemic uncertainty of out-of-distribution data  will be higher than that of in-distribution data:
\begin{equation}\label{eq_ubood}
  U_Q(\mathbb{O}) > U_Q(\mathbb{I})
\end{equation}

UBOOD directly uses the output of the epistemic uncertainty function $U_Q$ as the real-valued classification score.
As is the case for many one-class classificators, this real-valued score forms the input of a threshold-based decision function, which then assigns the in- or out-of-distribution class label.

\subsection{Classification Threshold}\label{subsec_class_threshold}
As is the case for any score-based one-class classification method, the classification threshold can be adjusted to modify the behaviour of the classifier, depending on the application's requirements.
For many applications, where some amount of OOD data is intermixed with the training data and the percentage is known, this information can be used to specify the threshold.
As in our case, per definition, there are no OOD samples in the training data, such an approach is not possible.
As a viable first solution, we propose the following simple algorithm to calculate a dynamic classification threshold:

\begin{enumerate}
  \item Calculate the average uncertainty of the in-distribution samples $\overline{U_{Q}} = \frac{1}{|\mathbb{I}|} \sum_{(s, a) \in \mathbb{I}}^{} U_Q(s, a)$.
  \item Treat $U_Q$ as a probability distribution and define the classification threshold as $c = \overline{U_Q} + \sigma(U_Q)$.
\end{enumerate}

\noindent Thus, a dynamic threshold-based on the uncertainty distribution is realized that adjusts over the training process as more data is gathered.
Please note that more complex algorithms for the threshold determination can be developed, e.g. by using multimodal probability distributions to model $U_Q$ 
or by making use of additional information about the available data on a per-application basis.

\subsection{Epistemic Uncertainty Estimation Methods}\label{subsec_epis_unc_est_approaches}
In principle, any of the epistemic uncertainty estimation methods mentioned in Section~\ref{subsec:uncertainty_in_dl} 
that are applicable to the function approximator used to model the Q-function, can be used in the UBOOD framework.
In this paper, we evaluate three different UBOOD versions using different methods for epistemic uncertainty estimation and their effect on the OOD classification performance,
as the networks are being used by the RL agent for value estimation.

\begin{figure}[ht]
  \centering
  \begin{subfigure}{.23\textwidth}
    \frame{\includegraphics[width=.98\textwidth]{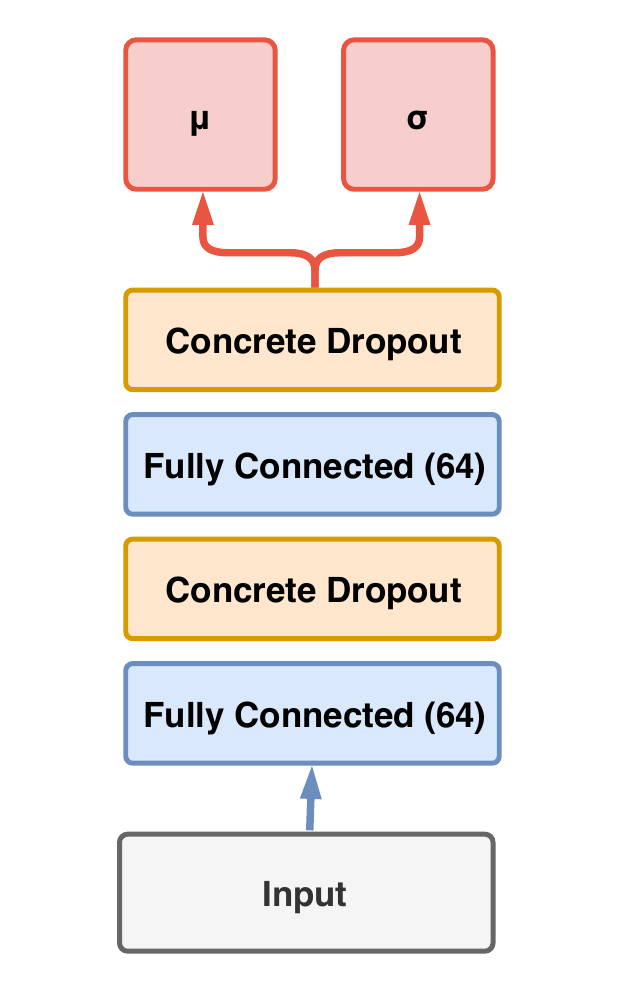}}
    \caption{MCCD network}
    \label{fig:models_mccd}
    \vspace*{3mm}
  \end{subfigure}
  \begin{subfigure}{.23\textwidth}
    \frame{\includegraphics[width=.98\textwidth]{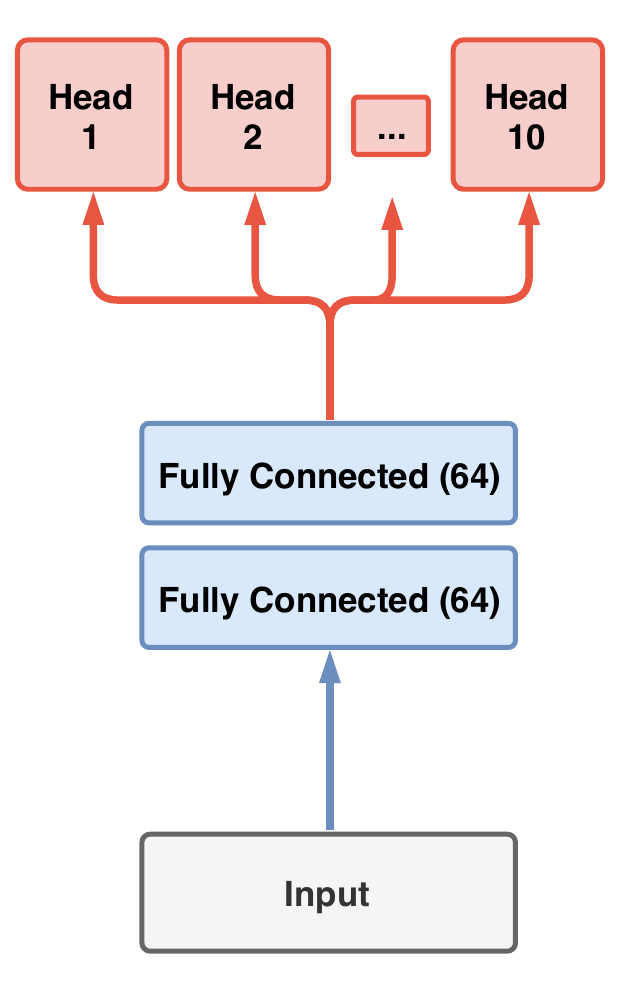}}
    \caption{Bootstrap network}
    \label{fig:models_boot}
    \vspace*{3mm}
  \end{subfigure}

  \begin{subfigure}{.46\textwidth}
    \frame{\includegraphics[width=.98\textwidth]{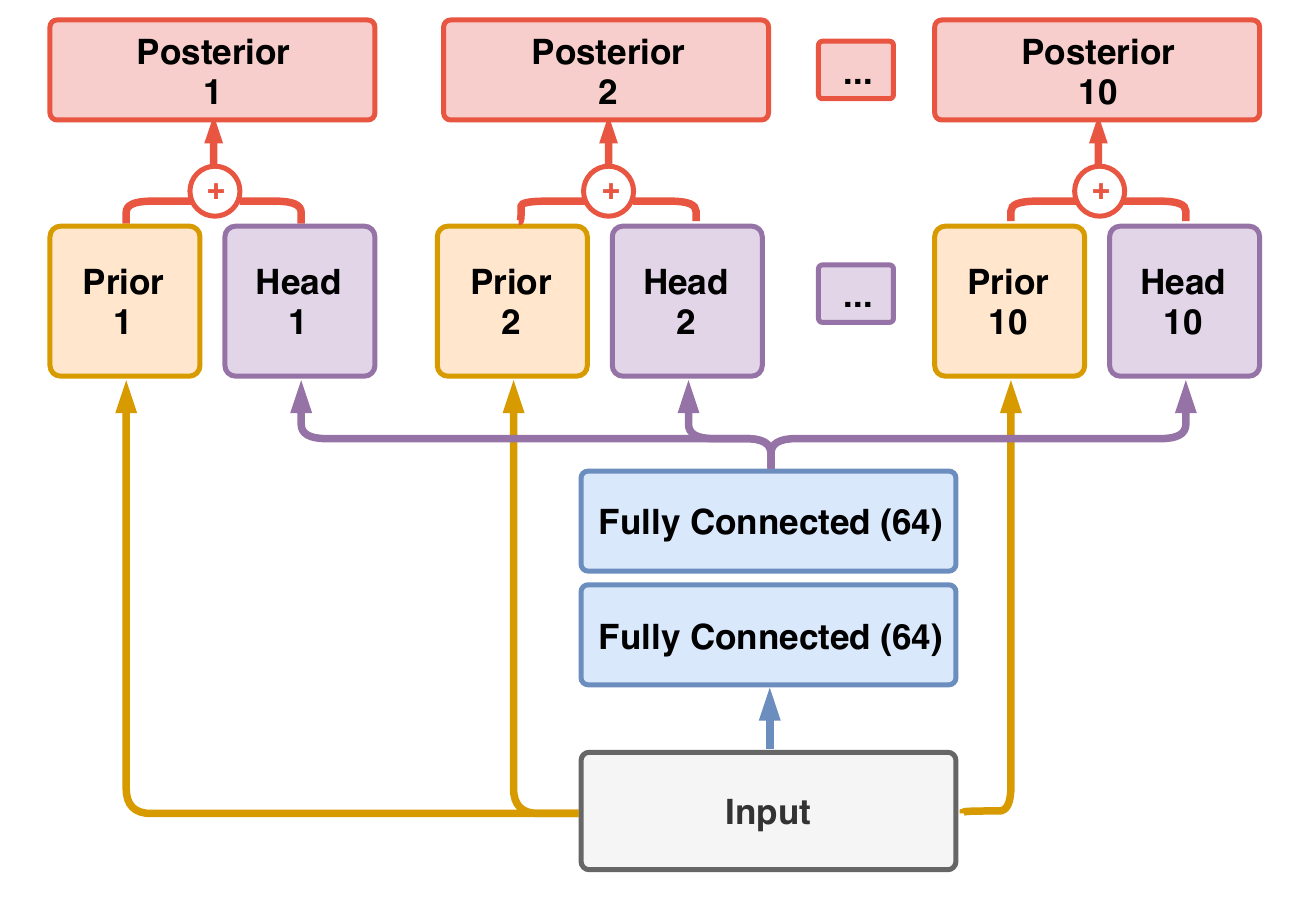}}
    \caption{Bootstrap-Prior network}
    \label{fig:models_bootp}
    \vspace*{2mm}
  \end{subfigure}
  \caption{Model architectures of the evaluated networks. (\protect\subref{fig:models_mccd}) The Monte-Carlo Concrete Dropout network. For this architecture, multiple MC samples are required to calculate the epistemic uncertainty. (\protect\subref{fig:models_boot}) The Bootstrap neural network with $K=10$ bootstrap heads, and (\protect\subref{fig:models_bootp}) the Bootstrap-Prior neural network which adds the output of an untrainable prior network to the output of the bootstrap heads to generate $K=10$ posterior heads. For both bootstrap-based architectures, epistemic uncertainty is calculated as the variance of the $K$ output heads.}
  \label{fig:models}
\end{figure}

The Monte-Carlo Concrete Dropout method is based on the dropout variational inference architecture as described by \cite{whatUncertainties17}.
Instead of default dropout layers, we use concrete dropout layers as described by \cite{galConcrete17},
which do not require pre-specified dropout rates and instead learn individual dropout rates per layer.
Figure \ref{fig:models_mccd} presents a schematic of the network used by this method. 

\noindent This concrete dropout method is of special interest in our context of reinforcement learning, as here the available data change during the training process,
rendering a manual optimization of the dropout rate hyperparameter even more difficult.
Model loss is calculated by minimizing the negative log-likelihood of the predicted output distribution.
Epistemic uncertainty as part of the total predictive uncertainty is then calculated as:
\begin{equation}\label{eq_epis_unc_dropout}
  \textrm{Var}_{ep}(y) \approx \frac{1}{T}\sum_{t=1}^{T}{\hat{y}_t^2} - (\frac{1}{T}\sum_{t=1}^{T}\hat{y}_t)^2
\end{equation}
with $T$ outputs $\hat{y}_t$ of the Monte-Carlo sampling.

The Bootstrap method is based on the network architecture described by \cite{bootstrappedDQN16}.
It represents an efficient implementation of the bootstrap principle by sharing a set of hidden layers between all members of the ensemble.
In the network, the shared, fully-connected hidden layers are followed by an output layer of size $K$, called the bootstrap heads, as can be seen in Figure \ref{fig:models_boot}.
For each datapoint, a Boolean mask of length equal to the number of heads is generated, which determines the heads this datapoint is visible to.
The mask's values are set by drawing $K$ times from a masking distribution.
For the work at hand, the  values are independently drawn from Bernoulli distributions with either $p=0.7$ or $p=1.0$.
In the case of $p=1.0$, the bootstrap is reduced to a classic ensemble where all heads are trained on the complete data.

The Bootstrap-Prior method is based on the extension presented in \cite{osband-prior18}.
It has the same basic architecture as the Bootstrap method but with the addition of a so-called random \textit{Prior Network}.
Predictions are generated by adding the data dependent output of this untrainable prior network to the output of the different bootstrap heads in order to calculate the ensemble posterior (Figure \ref{fig:models_bootp}). 
The authors conjecture that the addition of this randomized prior function outperforms deep ensemble-based methods without explicit priors,
as for the latter, the initial weights have to act both as prior and training initializer.

For both bootstrap-based methods, epistemic uncertainty is calculated as the variance of the $K$ outputs.

\section{\uppercase{Experimental Setup}}

\subsection{Framework versions}

We evaluate three different versions of the UBOOD framework:

\begin{itemize}
  \item \textbf{UB-MC}: UBOOD with Monte-Carlo Concrete Dropout (MCCD) network
  \item \textbf{UB-B}: UBOOD with Bootstrap network
  \item \textbf{UB-BP}: UBOOD with Bootstrap-Prior network
\end{itemize}

\noindent The UB-MC version's estimator network consists of two fully-connected hidden layers with 64 neurons each,
followed by two separate neurons in the output layer representing $\mu$ and $\sigma$ of a normal distribution.
As concrete dropout layers are used, no dropout probability has to be specified.
Model loss and epistemic uncertainty are calculated as described in Section~\ref{sec:u-bood}.

The UB-B Bootstrap neural network and UB-BP Bootstrap-Prior neural network versions all consist of two fully-connected hidden layers with 64 neurons each,
which are shared between all heads, followed by an output layer of $K=10$ bootstrap heads.

Each of these UBOOD versions is further evaluated with two parametrizations of the respective epistemic uncertainty estimation method:
UB-MC40 and UB-MC80 differ in respect to the amount of Monte-Carlo forward passes that are executed to approximate the epistemic uncertainty: $40$ or $80$ passes.
UB-B and UB-BP parametrizations (UB-B07, UB-B10, UB-BP07, UB-BP10) differ in respect to the Bernoulli distribution used to determine the bootstrap mask:
probability $p=0.7$ for UB-B07 \& UB-BP07 and probability $p=1.0$ for UB-B10 \& UB-BP10.

For all networks, ReLU is used as the layers' activation function, with the exception of the output layers, where no activation function is used.
The classification threshold is calculated as $c = \overline{U_Q} + \sigma(U_Q)$, as described in section \ref{subsec_class_threshold}.

\subsection{Environments}
\label{subsec:envs}
One of the problems in evaluating OOD detection for RL is the lack of datasets or environments
which can be used for generating and assessing OOD samples in a controlled and reproducible way.
By contrast to the field of image classification, where benchmark datasets like \textit{notMNIST} \cite{notmnist11} exist that contain OOD samples,
there are no equivalent sets for RL.
We apply a principled approach to develop two environments, one using a gridworld-style discrete state-space, the other using a continuous state-space.
Both environments allow systematic modifications after the training process, thus producing OOD states during evaluation.

\begin{figure}[ht]
  \centering
  \begin{subfigure}{.49\textwidth}
    \includegraphics[width=.98\textwidth]{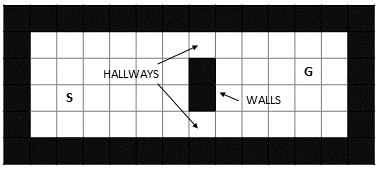}
    \caption{Example environment: Config 0}
    \label{fig:factory_env0}
    \vspace*{2mm}
  \end{subfigure}
  
  \begin{subfigure}{.49\textwidth}
    \includegraphics[width=.98\textwidth]{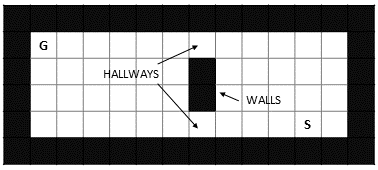}
    \caption{Example environment: Config 7}
    \label{fig:factory_env7}
    \vspace*{2mm}
  \end{subfigure}
  \caption{Example initializations of the gridworld pathfinding environment using different configurations. The label \textit{S} indicates the agent's start position, while \textit{G} marks the goal. Both positions are randomly set in the ranges defined by the respective configuration every episode. (\protect\subref{fig:factory_env0}) shows a placement using environment configuration 0 as active in training. Samples collected with this configuration define the in-distribution set. (\protect\subref{fig:factory_env7}) shows an initialization of environment configuration 7 which differs maximally from the training configuration.}
  \label{fig:factory_env}
\end{figure}

The first environment is a simple gridworld pathfinding environment.
It is built on the design presented in \cite{sedlmeier2019uncertainty} and has a discrete state-space.
The basic layout consists of two rooms, separated by a vertical wall.
Movement between the rooms is only possible via two hallways, as is visualised in Figure \ref{fig:factory_env}.
The agent starts every episode at a random position on the grid (labeled \textit{S} in Figure \ref{fig:factory_env}).
Its task is to reach a specific goal position on the grid (labeled \textit{G} in Figure \ref{fig:factory_env}),
which also varies randomly every episode, by choosing one of the four possible actions: $\{\textit{up,down,left,right}\}$.

The state of the environment is represented as a stack of three $12 \times 4$ feature planes,
with each plane representing the spatial positions of all environment objects of a specific type: agent, goal or wall.
Each step of the agent incurs a cost of $-1$ except the goal-reaching action, which is rewarded with $+100$ and ends the episode.
We evaluate the performance of the UBOOD framework on a set of $8$ environment configurations.
All environment configurations have a size of $12 \times 4$ and randomly vary the y-coordinate of the agent's start position as well as the goal position every episode, in the interval $[0,4)$.
Configuration $0$, the only configuration used in training, varies the x-coordinate of the agent's start position in the interval $[0,5)$ and the goal position in the interval $[7,12)$.
Each environment configuration $1-7$ is then defined by shifting the start interval right by $1$ compared to the previous configuration, while the goal interval is shifted left by $1$. E.g. configuration $1$ has start position range $[1,6)$ and goal position range $[6,11)$.
This results in environment configurations with increasing difference from the training configuration $0$, as can be seen in the example shown in Figure \ref{fig:factory_env7}.

\begin{figure}[t]
  \centering
  \begin{subfigure}{.24\textwidth}
    \includegraphics[width=.98\textwidth]{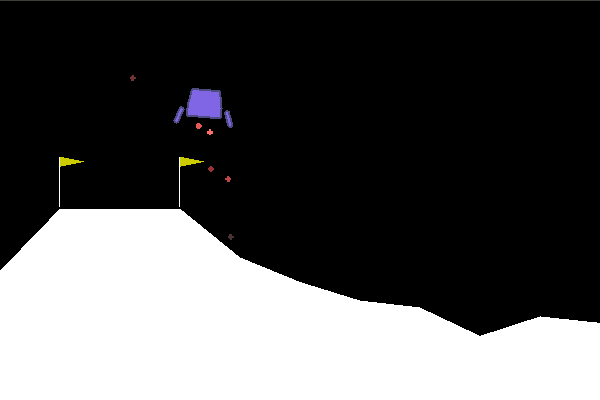}
    \caption{Example: Config 0}
    \label{fig:lunar_env0}
  \end{subfigure}%
  \begin{subfigure}{.24\textwidth}
    \includegraphics[width=.98\textwidth]{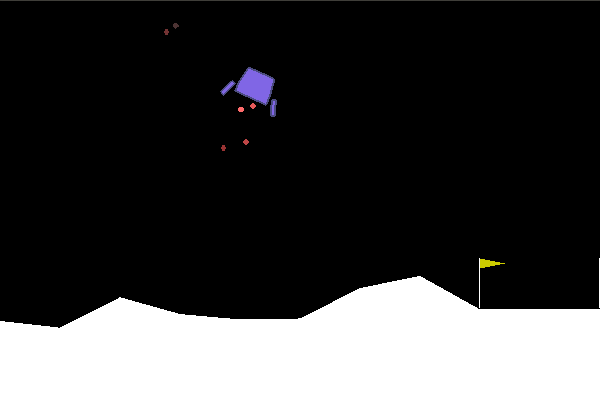}
    \caption{Example: Config 5}
    \label{fig:lunar_env5}
  \end{subfigure}
  \vspace*{2mm}
  \caption{Examples from the LunarLander environment using different configurations. (\protect\subref{fig:lunar_env0}) Example using environment configuration 0 as active in training. Samples collected with this configuration define the in-distribution set. Example using (\protect\subref{fig:lunar_env5}) environment configuration 5 which differs maximally from the training configuration.}
\end{figure}

The continuous state-space environment is based on OpenAI's LunarLander environment \cite{openai2016gym}.
The goal is to safely land a rocket inside a defined landing pad, without crashing. This task can be understood as rocket trajectory optimization.
While the original environment defines a static position for the landing pad, our modified environment allows for random placement inside specified intervals. 
As the original environment does not encode the landing pad's position in the state representation, our version extends the state encoding to include the left and right x-coordinate as well as the y-coordinate of the pad.
For evaluating the performance of the UBOOD framework in this continuous state-space environment, we created a set of $6$ configurations.
Configuration $0$, the only configuration used in training, varies the x-coordinate of the center of the landing pad in the interval $[2,5)$ and the y-coordinate in the interval $[6,12)$, which results in the landing pad being placed in the upper left side of the environment. An example of this configuration can be seen in Figure~\ref{fig:lunar_env0}.
Each environment configuration $1-5$ is then defined by shifting the x-coordinate interval right by $1$ compared to the previous configuration, while the y-coordinate interval is shifted left by $1$. This results in the pads being placed increasingly to the lower right side of the environment.
Like in the gridworld environment, this produces environment configurations with increasing difference from the training configuration $0$.

Note that training on both environments is solely performed using the respective environment configuration $0$.
Evaluation runs are executed independently of the training process, based on model snapshots generated at the respective training episodes.
Consequently, data collected during these evaluation runs is not used for training.

\section{\uppercase{Performance Results}}

\begin{figure}[t]
  \includegraphics[width=.48\textwidth]{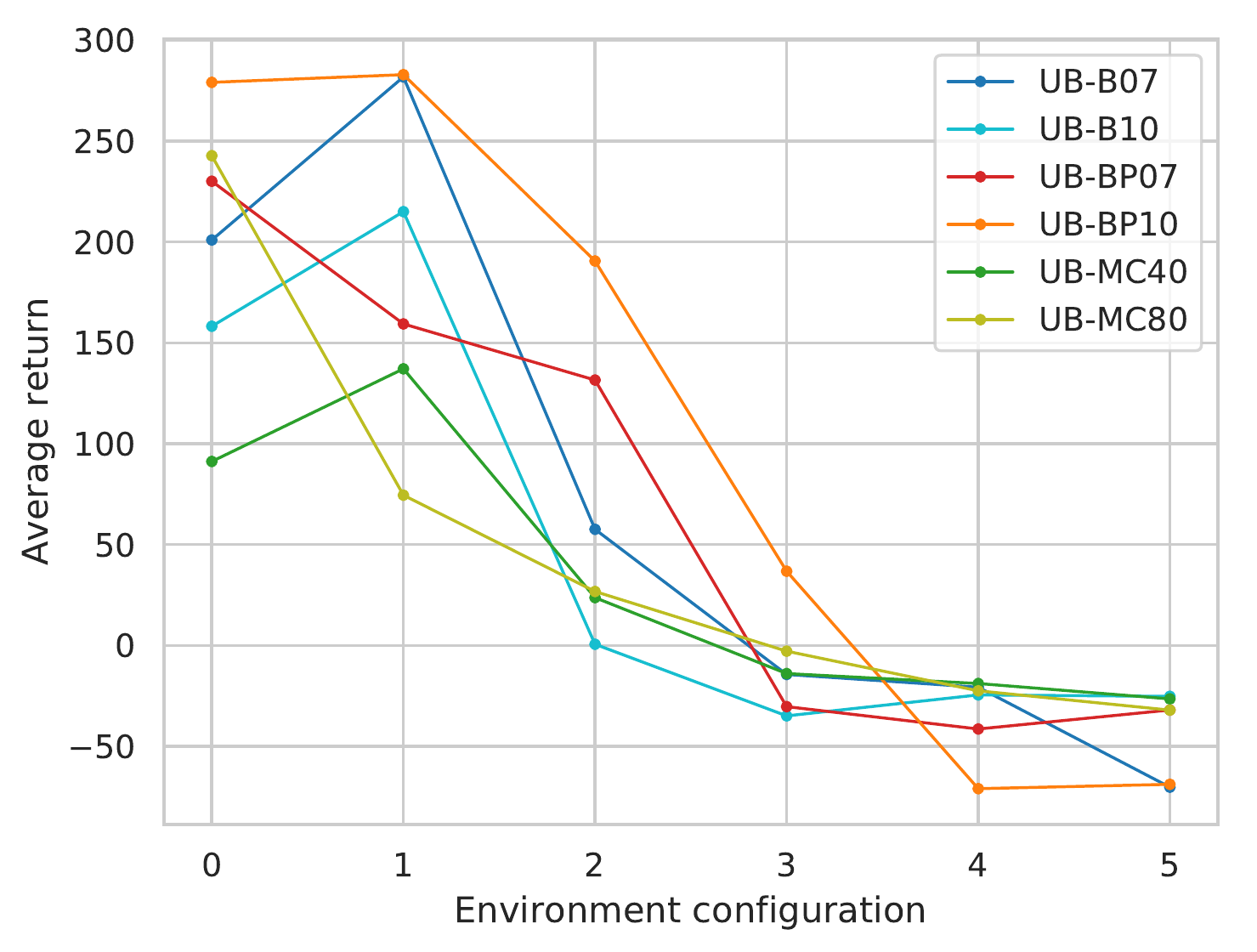}
  \caption{Returns achieved by the different versions on varying configurations of the LunarLander environment after $10000$ training episodes on configuration $0$. Envionment configurations $1-5$ modify the environment with increasing strength as described in Section~\ref{subsec:envs}. All values shown are averages of $30$ evaluation runs.}
  \label{fig:lunar_return}
\end{figure}

\noindent All evaluated versions learn successful policies on both the gridworld and LunarLander environments.
Returns achieved by the trained policies after $10000$ training episodes on different environment configurations are shown in Figure~\ref{fig:lunar_return}.
As is to be expected, increasing changes to the environment (configuration $1-5$) reduce the achieved return, as the evaluation environment increasingly differs from the training environment configuration $0$.

\begin{figure}[ht]
  \centering
  \begin{subfigure}{.48\textwidth}
    \includegraphics[width=.99\textwidth]{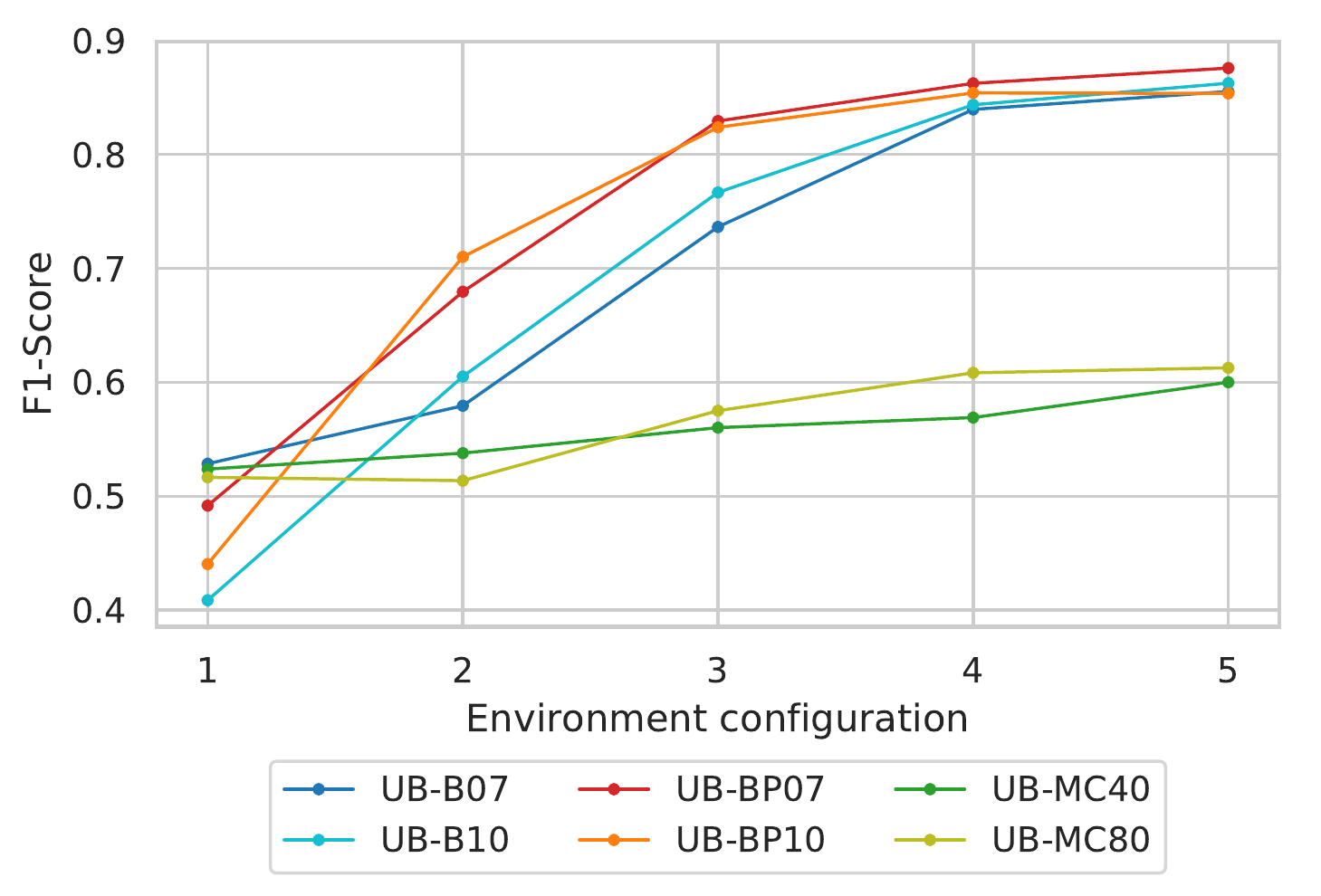}
    \label{fig:lunar_f1}
    \caption{LunarLander}
  \end{subfigure}
  \begin{subfigure}{.48\textwidth}
    \includegraphics[width=.99\textwidth]{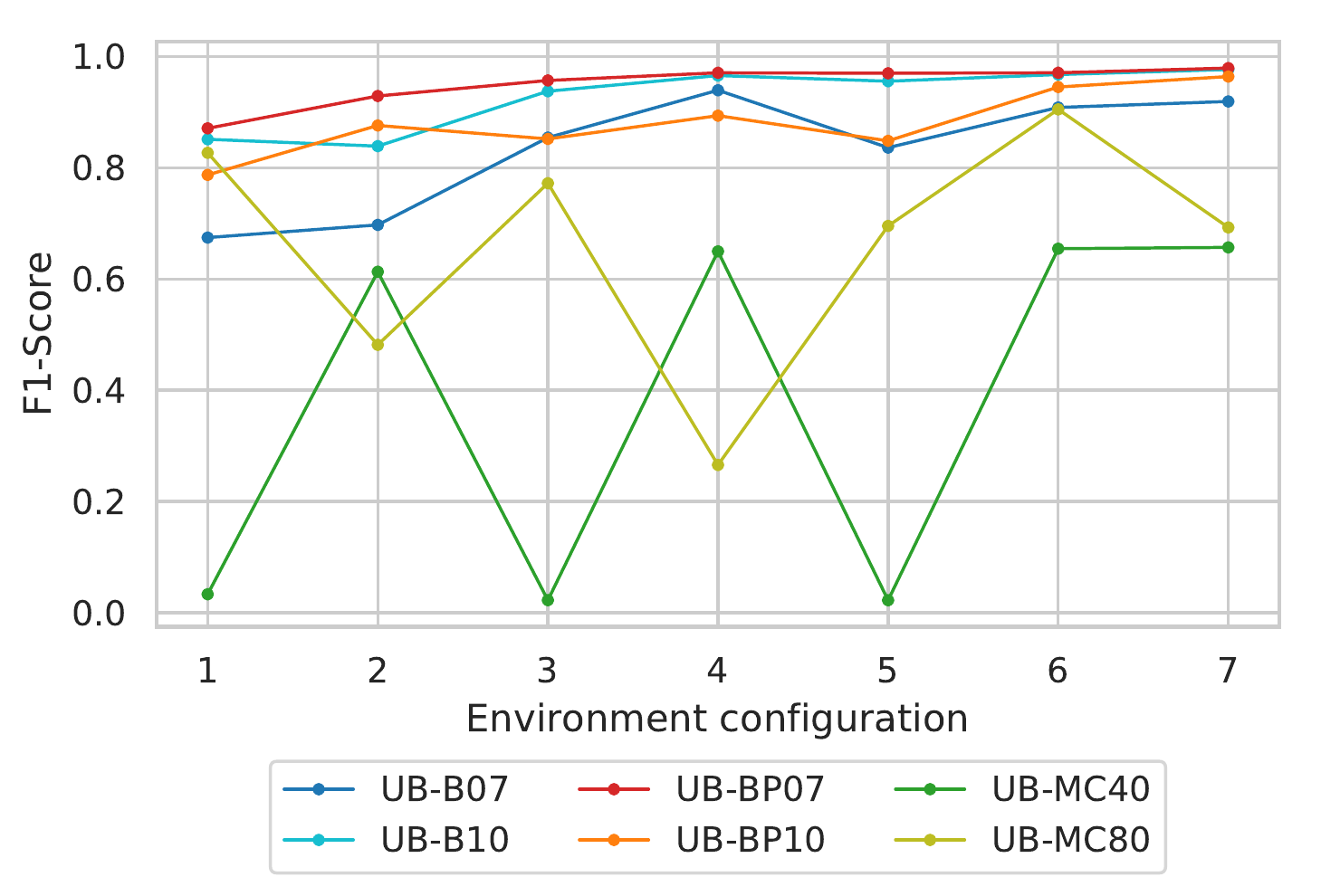}
    \label{fig:gridworld_f1}
    \caption{Gridworld}
  \end{subfigure}
  \caption{F1-Scores of the classifier evaluated on different configurations of the LunarLander and gridworld environments. Samples collected on the training configuration of each environment are defined as \textit{negatives} (in-distribution), samples from the other configurations $1-5$ as \textit{positives} (OOD). X-Axis shows evaluations performed with samples from the training configuration and the respective environment configuration $1-5$. Samples are aggregated from $30$ consecutive episode runs.} 
  \label{fig:f1-scores}
\end{figure}

We evaluate the performance of the UBOOD framework based on the F1-Score as the harmonic mean of precision and recall.
Figure~\ref{fig:f1-scores} shows the F1-Scores achieved, dependent on the uncertainty estimation technique used in the framework.
Best overall classification results on the LunarLander environment are achieved for UB-BP, i.e. using UBOOD with the Bootstrap-Prior estimator with F1-values as high as $0.903$ for UB-BP07 on environment configuration $5$.
F1-Scores of the UB-B and UB-BP versions on the gridworld environment are higher overall, when compared to the UB-MC versions.
Here, values range between a minimum of $0.674$ on evaluation configuration $1$, which is closest to the training configuration,
and $0.958$ on configuration $5$, which produces the strongest OOD samples.
Overall, classification performance increases over environment configurations $1-5$ when Bootstrap-based estimators are used in the UBOOD framework.
UB-MC, i.e. UBOOD combined with MCCD estimators, generates highly varying F1-scores, ranging between $0.020$ and $0.738$ on the gridworld environment
and $0.280$ and $0.484$ on the LunarLander environment.
By contrast to the Bootstrap-based versions, there is no relation apparent between the strength of the environment modification and the classification performance.

\begin{figure}[t]
  \includegraphics[width=.49\textwidth]{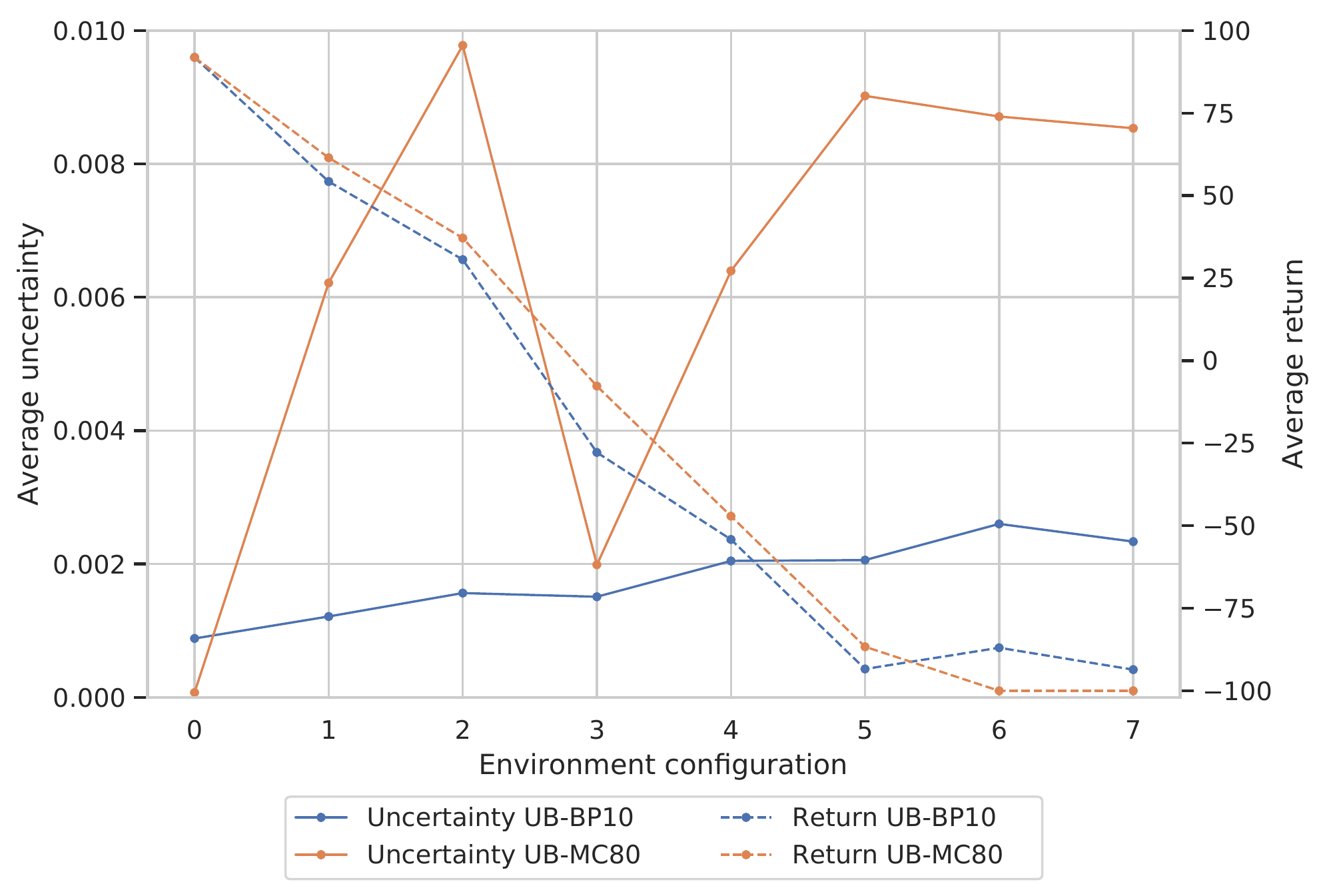}
  \caption{Uncertainty VS return of UB-BP10 and UB-MC80 evaluated on different configurations of the gridworld environment. While for the Bootstrap-based version UB-BP10, increases in uncertainty are reflected in decreases of return, a large decrease in uncertainty is visible for UB-MC80 between configuration $2$ and $3$, although the achieved return also decreases. All values shown are averages of $30$ evaluation runs.}
  \label{fig:uncertainty_vs_reward}
\end{figure}

We further evaluate the relation between reported uncertainty and the return achieved by the agent.
Figure~\ref{fig:uncertainty_vs_reward} shows evaluation results of the UB-BP10 and UB-MC80 versions evaluated on different configurations of the gridworld environment.
For UB-BP10 ($p=1.0$), increases in uncertainty (caused by increasing environment modifications) are reflected in decreases of return.
This behaviour was also present on the LunarLander environment and consistent for different values of $p$.
No such clear relation was visible for UB-MC80.
As can be seen in the results visualised in Figure~\ref{fig:uncertainty_vs_reward}, the uncertainty reported by the MCCD-based version decreases strongly between configuration $2$ and $3$, although the achieved return also decreases.

\section{\uppercase{Discussion and Future Work}}

\noindent In this paper, we proposed UBOOD, an uncertainty-based out-of-distribution classification framework.
Evaluation results show that using the epistemic uncertainty of the agent's value function presents a viable approach for OOD classification in a deep RL setting.
We find that the framework's performance is ultimately dependent on the reliability of the underlying uncertainty estimation method, which is why good uncertainty estimates are required.

\begin{figure}[t]
  \centering
  \begin{subfigure}{.47\textwidth}
    \includegraphics[width=.98\textwidth]{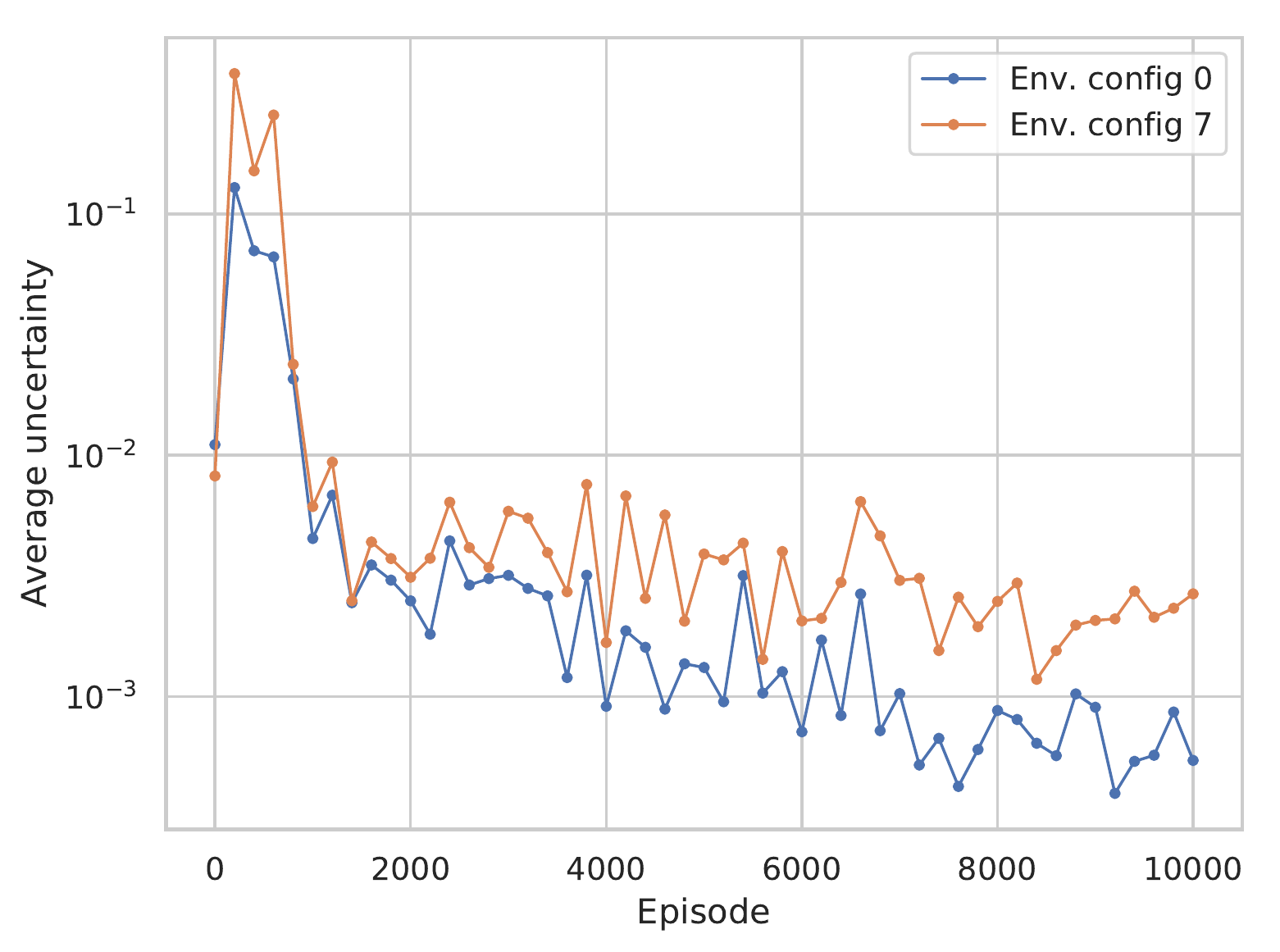}
    \caption{UB-B07}
    \label{fig:factory_uncertainty_over_ep_boot}
  \end{subfigure}
  \begin{subfigure}{.47\textwidth}
    \includegraphics[width=.98\textwidth]{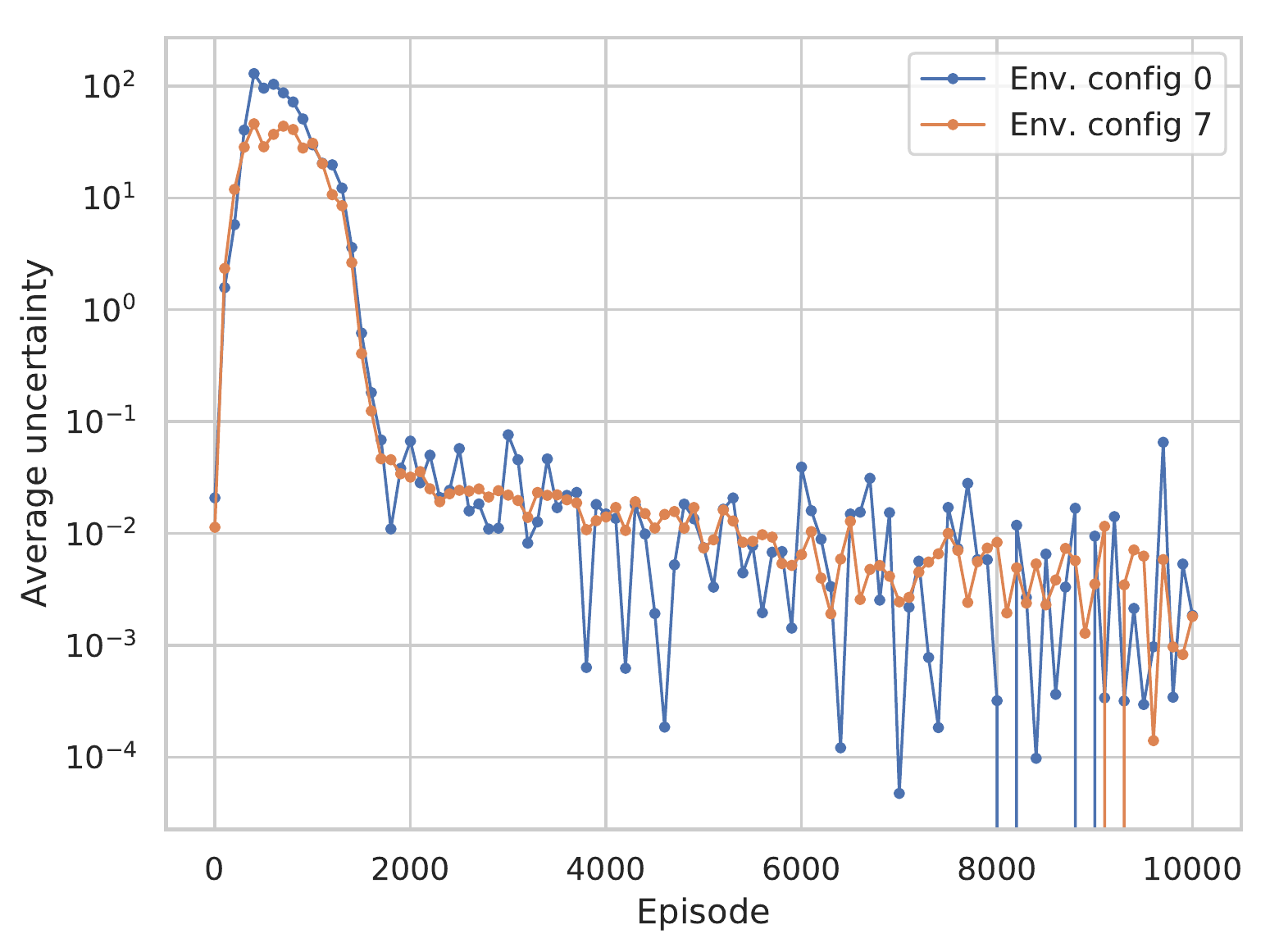}
    \caption{UB-MC80}
    \label{fig:factory_uncertainty_over_ep_mccd}
  \end{subfigure}
  \vspace*{2mm}
  \caption{Average uncertainties reported by (\protect\subref{fig:factory_uncertainty_over_ep_boot}) Bootstrap-based version UB-B07 and (\subref{fig:factory_uncertainty_over_ep_mccd}) Monte-Carlo Concrete Dropout based version UB-MC80 on the Gridworld environment. \textit{Env. config 0} shows uncertainties reported on the training configuration of the environment (in-distribution), \textit{Env. config 7} the uncertainties on the maximaly diverging configuration. While for UB-B07 the uncertainties start diverging with progressing training, there is no such effect for UB-MC80. As a consequence, only the Bootstrap-based version allows for an increasingly better differentiation between in- and OOD samples. All values shown are averages of $30$ evaluation runs.}
  \label{fig:factory_uncertainty_over_ep}
\end{figure}

On both evaluation domains, UBOOD combined with ensemble-based bootstrap uncertainty estimation methods (UB-B / UB-BP) shows good results with F1-scores as high as $0.903$, allowing for a reliable differentiation between in- and OOD-samples.
F1-Scores increase as the environment configuration differs more from the training environment, i.e. the stronger OOD the observed samples, the more reliable the classification.
The addition of a prior as done with the UB-BP version seems to have a positive effect on the separation of in- and out-of-distribution samples as is reflected in higher F1-scores on the LunarLander environment.
By contrast, UBOOD combined with the concrete dropout-based uncertainty estimation method (UB-MC) does not produce viable results.
Although increasing the amount of Monte-Carlo samples improves the performance somewhat, the resulting classification performance is not on par with the Bootstrap-based versions.
The reason for the large difference in performance between the Bootstrap-based and MCCD-based versions can be seen in the example shown in Figure ~\ref{fig:factory_uncertainty_over_ep}.
For the UB-B version, the reported uncertainties on environment configuration $0$ (training) and $7$ (strong modification) increasingly diverge with progressing training episodes (Figure~\ref{fig:factory_uncertainty_over_ep_boot}).
As this is not the case for the UB-MC version (Figure~\ref{fig:factory_uncertainty_over_ep_mccd}), only the Bootstrap-based version allows for an increasingly better differentiation between in- and OOD samples and consequently high F1-scores of the classifier.
We found this effect to be consistent over all parametrizations of the Bootstrap- and MCCD-based versions we evaluated.

Our results match recent findings \cite{Beluch_2018_CVPR}, where ensemble-based uncertainty estimators were compared against Monte-Carlo Dropout based ones for the case of active learning in image classification.
Results presented in that work also showed that ensembles performed better and led to more calibrated uncertainty estimates.
As a possible explanation, the authors argue that the difference in performance is a result of a combination of decreased model capacity
and lower diversity of the Monte-Carlo Dropout methods when compared to ensemble approaches.
This effect would also explain the behaviour we observed when comparing reported uncertainty and achieved return.
While there is a strong inverse relation visible when using Bootstrap-based UBOOD versions, no clear pattern emerged for the evaluated MCCD-based versions. 
We think that further research into the relation between epistemic uncertainty and achieved return when train- and test-environments differ could provide interesting insights relating to generalization performance in deep RL.
Being able to differentiate between an agent having encountered a situation in training versus the agent generalizing its experience to new situations could provide a huge benefit in safety-critical situations.

\bibliographystyle{apalike}
{\small
  \bibliography{main}}

\end{document}